\documentclass{article}

\usepackage{arxiv}

\usepackage[utf8]{inputenc} 
\usepackage[T1]{fontenc}    
\usepackage{hyperref}       
\usepackage{url}            
\usepackage{booktabs}       
\usepackage{amsfonts}       
\usepackage{nicefrac}       
\usepackage{microtype}      
\usepackage{lipsum}
\usepackage[pdftex]{graphicx}
\usepackage{listings}
\usepackage{float}

\title{ANNdotNET - deep learning tool on .NET Platform}

\author{
  Bahrudin I.~Hrnjica\thanks{ Personal web page: \url{http://bhrnjica.net}, GitHub: \url{http://github.com/bhrnjica}.} \\
  Faculty of Engineering Sciences\\
  University of Bihac\\
  77000, Bihac\\
  Bosnia and Herzegovina \\
  \texttt{bahrudin-hrnjica@unbi.ba} \\
}

\begin{document}
\maketitle

\begin{abstract}
ANNdotNET – is an open source project for deep learning written in C\# with ability to create, train, evaluate and export deep learning models. The project consists of the Graphical User Interface module capable to visually prepare data, fine tune hyper-parameters, design network architecture, evaluate and test trained models. The ANNdotNET introduces the Visual Network Designer, (VND) for visually design almost any sequential deep learning network. Beside VND, ANNdotNET implements Machine Learning Engine, (MLE) based on CNTK - deep learning framework, with ability to train and evaluate models on GPU. For model evaluation ANNdotNET contains rich set of visual and descriptive performance parameters, history of the training process and set of export/deployment options. The advantage of using ANNdotNET over the classic code based ML approach is more focus on deep learning network design and training process instead of focusing on coding and debugging. It is ideal for engineers not familiar with supported programming languages. The project is hosted at \url{http://github.com/bhrnjica/anndotnet}.
\end{abstract}


\section{Introduction}

ANNdotNET – is .NET based solution consisting of set of tools for running deep learning models. The process of creating, training, evaluating and exporting models is provided by the GUI based Application and does not require knowledge for supported programming language. The ANNdotNET GUI Tool implements functionalities for data preparation prior to training process. The module consists of functionalities for data cleaning, feature selection, category encoding, missing values handling, creation of training and validation set. Once the data is prepared, the user can create empty DL model to start building, training and evaluate it. 

ANNdotNET introduces the Visual Network Designer, VND for visually design deep neural networks. Design process is completely visual and no coding is required. It helps the user to focus on deep network design rather than debugging the code. VND supports the most popular and widely used network layers such as Dense, LSTM, Convolutional,Pooling, DropOut, etc. Also VND can be used in order to design more complex layers such as AutoEncoders, Embedding, etc. 

ANNdotNET introduces the ANNdotNET Machine Learning Engine (MLE) which is responsible for training and evaluation of DL models. The MLE relies on Microsoft Cognitive Toolkit (CNTK) open source library developed by Microsoft\cite{cntk2014}. 

For evaluation and test of the trained DL models, ANNdotNET provides set of visually presented performance parameters that can be used for regression, binary and multi-class classification models, history of the training process, early stopping, etc.

Information collected during DL creation process are stored in the set of hierarchically organized files. In ANNdotNET stores information into several different file types such as: project file (\emph{*.ann}), mlconfing file (\emph{*.mlconfig}), data file (\emph{*.txt}), history file (\emph{*.history}). Each file stores different kind of information important for the ML project.

\subsection{ANNdotNET key features}

As a desktop application ANNdotNET is suitable in several scenarios over the classic code based ML approaches:

\begin{itemize}
\item more focus on network development and training process using classic desktop approach, instead of focusing on coding,
\item less time spending on debugging source code, more focusing on different configuration and parameter variants,
\item fast development of deep learning network which can be quickly tested and implemented 
\item ideal for engineers/users who are not familiar with programming languages,
\item in case the problem requires more complex scenarios where additional coding implementation is required, the ANNdotNET provides high level API for such implementation,
\item all ML configurations developed with GUI tool,can be handled with command line based tool and vice versa.
\end{itemize}

\subsection{ANNdotNET Start Page}

In order to easy start working with, ANNdotNET comes with dozens of pre-calculated deep learning projects included in the installer. They can be accessed from the Start page. The pre-calculated projects are based on famous datasets from several categories: regression, binary and multi class classification problems, image classifications, times series, etc. In pre-calculated projects the user can find how to use various types of deep neural network configurations. Also, each pre-calculated project can be modified in terms of change its network configuration, learning and training parameters, as well as create new ml configurations based on the existing data set.

\begin{figure}[h]
\centering
\includegraphics[width=0.8\textwidth]{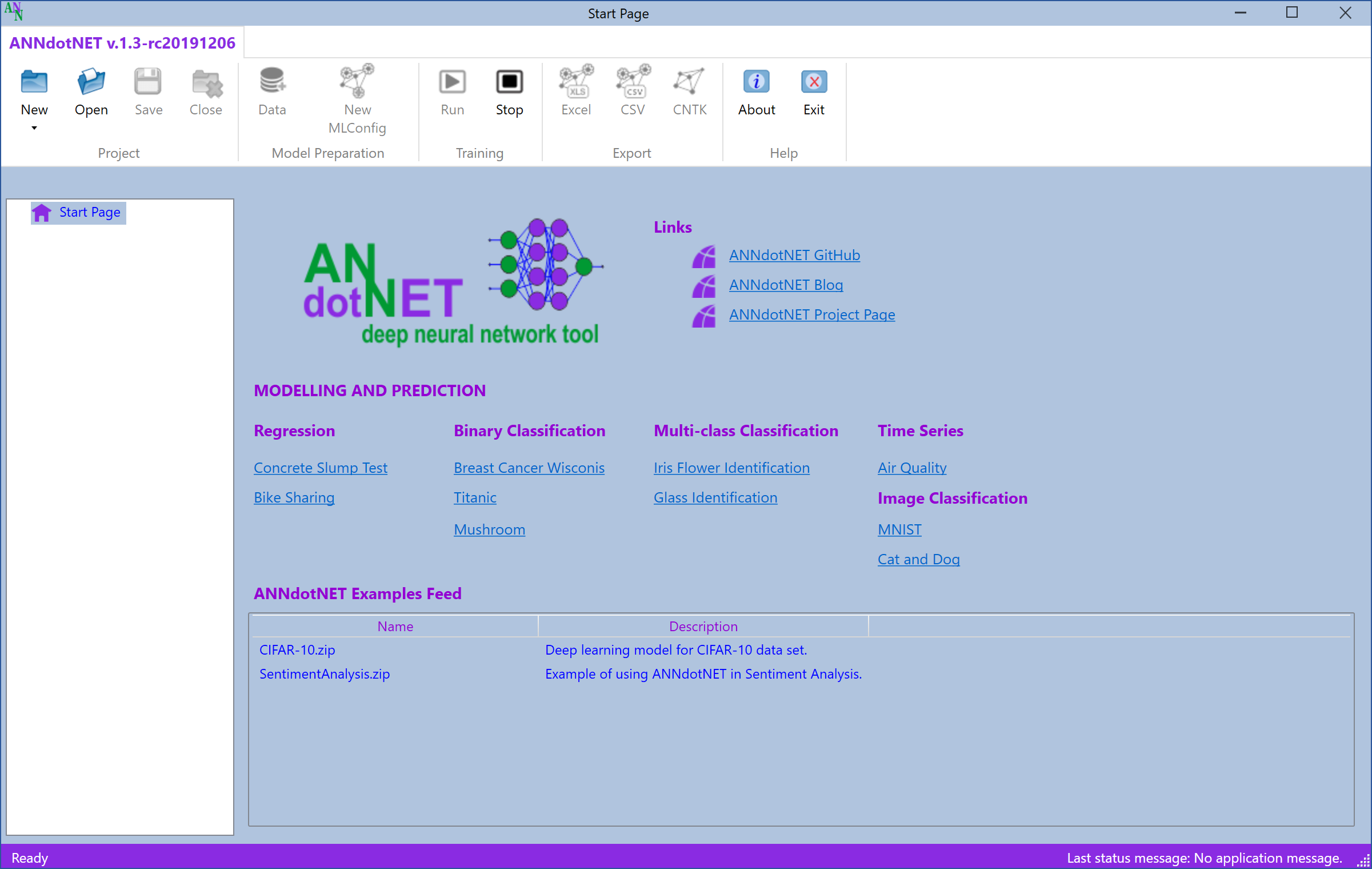}
\caption{ANNdotNET Start Window}
\label{fig_01} 
\end{figure}

The set of pre-calculated deep learning projects are not static. \emph{ANNdotNET Examples Feed} contains dynamic list of deep learning projects loaded from the GitHub repository. By Adding new deep learning project into the examples feed, every user running ANNdotNET can use it through the feed.

\section{Introduction of the project}

\subsection{Hardware requirements}

ANNdotNET support training and model evaluation on modern NVIDIA GPUs, however the training and evaluation can also be performed on CPU with older x64 processors with at least 2 GB of RAM. The minimal processors and memory requirements depends of training model.

\subsection{Software requirements}

In order to run and develop ANNdotNET based solution the following software requirements must be met:

\begin{itemize}
    \item Windows 8 x64 or higher,
    \item .NET Framework 4.7.2 and newer,
    \item .NET Core 2.0 and newer,
    \item Visual Studio 2019 (Community, Professional or Enterprise),
    \item Git source control tool.
\end{itemize}

In order to run and use GUI Tool for training deep learning models the machine requires the following software components:

\begin{itemize}
    \item Windows 8 x64 or higher,
    \item .NET Framework 4.7.2 and newer,
    \item .NET Core 2.0 and newer,
\end{itemize}

\subsection{Organization of the source code}

The ANNdotNET project is Visual Studio based solution consisted of several projects grouped into logical folders. In order to build the solution at least Visual Studio 2019 Community version should be installed on the local machine.

ANNdotNET solution can be grouped on several components:

\begin{itemize}
\item	The library 
\item	Command Tool 
\item	GUI Tool
\item	Excel AddIn
\item	Unit Tests and Test applications
\end{itemize}

\textbf{The library} consists of visual studio projects which logically separate the implementation. It provides foundation of data processing and preparation, neural network configuration and layers implementation, training and handling with minibatches. Within the library folder each project exposes set of API for the model evaluation, testing, export and deployment.

\textbf{Command Tool} is console-based tool which can be run from Visual Studio and can perform model training and evaluation using console output. 

\textbf{GUI Tool} is Windows desktop application which provides rich set of options and visualizations during machine learning steps: project and model creation, data preparation, model training, model evaluation and validation, export options and model deployment. 

\textbf{Excel AddIn } is implementation of Microsoft Office AddIn for model deployment into Excel. Using ANNdotNET Excel AddIn, trained model can be used in Excel like ordinary excel formula. This is very handy for model deployment into production when only Excel is need in order to use the model.

\textbf{Unit Tests} – set of unit tests and console projects for testing the implementation of the solution.

\section{GUI tool, projects, models and related files}

The basic object in ANNdotNET is machine learning configuration file, shortly named \emph{mlconfig}. The \emph{mlconfig}, with file extension $*.mlconfig$, holds information about features, labels, learning and training parameters, neural network architecture and set of paths required for training and evaluation, best trained model, training history etc. Simply said it is the representation of a deep learning model. Beside \emph{mlconfig} file ANNdotNET supports project file. The project file ($*.ann$) holds the information about whole ML project. It can consists of one or more \emph{mlconfig} files, data files and project info file.

\begin{figure}[h]
\centering
\includegraphics[width=0.8\textwidth]{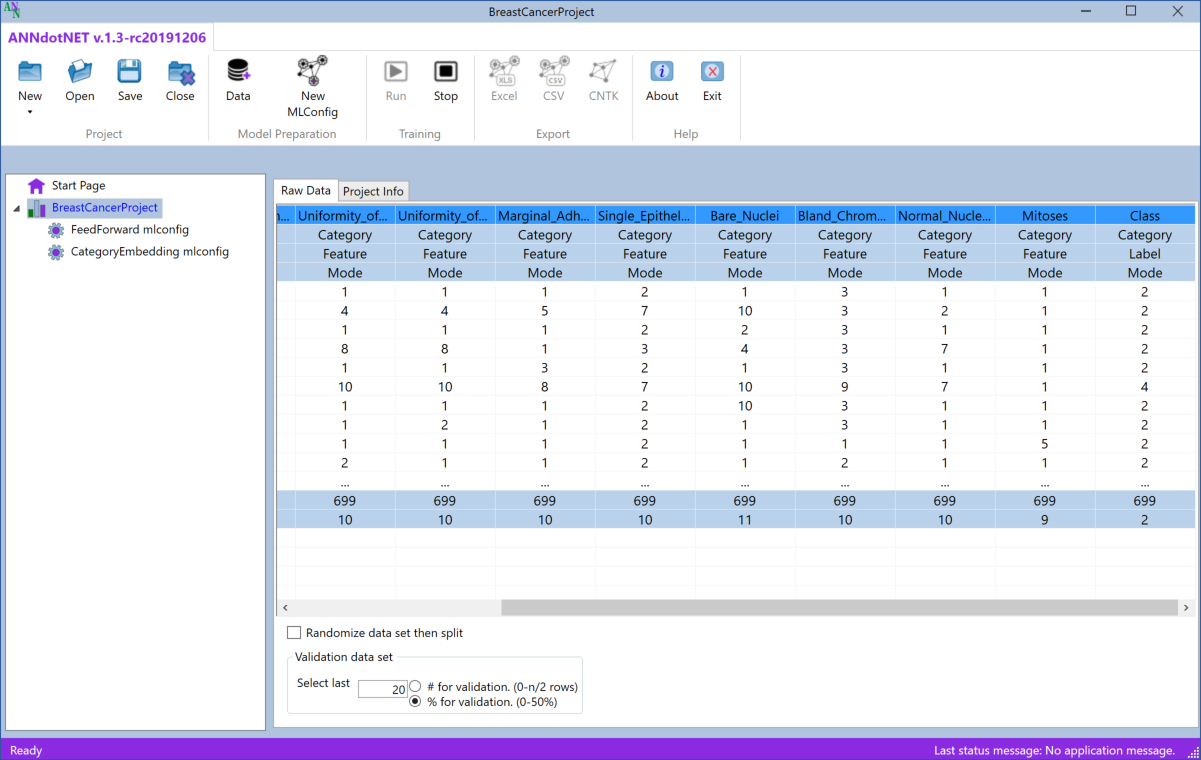}
\caption{ANNdotNET with opened Brest Cancer DL project. The Project explorer shows \emph{Breast Cancer} project with data set. Data is organized in columns. Each column is identified as \emph{features} or \emph{label}. Also for each column several additional information is defined such: column type, missing value handling and data normalization. }
\label{fig_02} 
\end{figure}

The user start working in ANNdotNET by creating new project. Then a data is loaded in order to start working on data preparation and feature selection. Once the project creation and data preparation are completed the new model ($mlcofing$ file) can be created. Example of a project $"Breast Cancer Project"$ with two models named: $FeedForward$ and $CategoryEmbedding$ are shown in Figure \ref{fig_02}. The project is based on famous Breast Cancer data set \cite{Wolberg1993}.

As can be seen the project is consisted of two DL models (two mlconfig files). Each model is created from different network architecture, different kind of training parameters and the same data set.  Figure \ref{fig_02} also show \emph{Project explorer} -tree control which shows a hierarchical representation of a project and related models. The user start with project creation, data loading and preparation and then can create as many models as necessary. 

\subsection{File structure in ANNdotNET}

While creating a new project the project file and project folder are created on disk. Illustration of a file and folder structure can be described as follow:
Assume one create a new project called $Project01$. The folder named $Projecet01$ is created, at the same time as  project file named $Project01.ann$. Those two items are shown on the following image:

\begin{figure}[h]
\centering
\includegraphics[width=15cm]{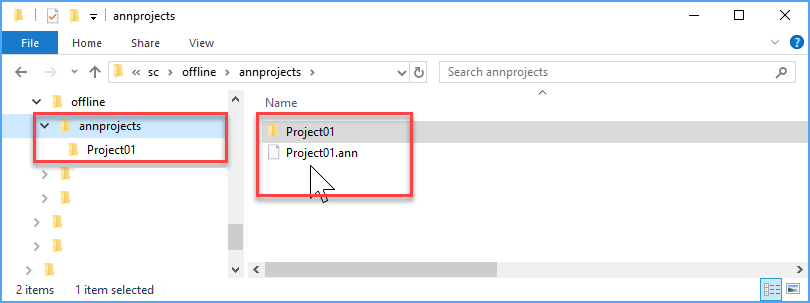}
\caption{Project file and folder structure in ANNdotNET}
\label{fig_03} 
\end{figure}

Once the project is created, one can load the data set file. The data set file is the file that contains data used for training and evaluation of the deep learning model. The structure of the data set is classic table-based textual data. For example one can load \url{https://archive.ics.uci.edu/ml/machine-learning-databases/iris/iris.data} file directly into ANNdotNET and start processing the data in order to implement deep learning model. Once the data is loaded, ANNdotNET processes the file and saved the copy of the data into the root of the project folder. 

\begin{figure}[h]
\centering
\includegraphics[width=15cm]{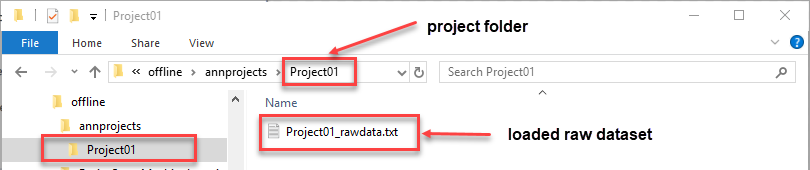}
\caption{Raw data set file within the project folder}
\label{fig_04} 
\end{figure}

During  data set file processing the new file is created in the project folder and named according to the ANNdotNET naming convention e.g. $[ProjectName]\_rawdata.txt$

Now that the project has been created and data set have loaded and processed the next step is to start building DL model. Each time the new DL model is created a coresponded \emph{mlfoncig} file is created on disk. Within a project there can be created arbitrary number of DL models with different structure and size of training and validation data sets and also with different network, learning and training parameters.

\begin{figure}[h]
\centering
\includegraphics[width=15cm]{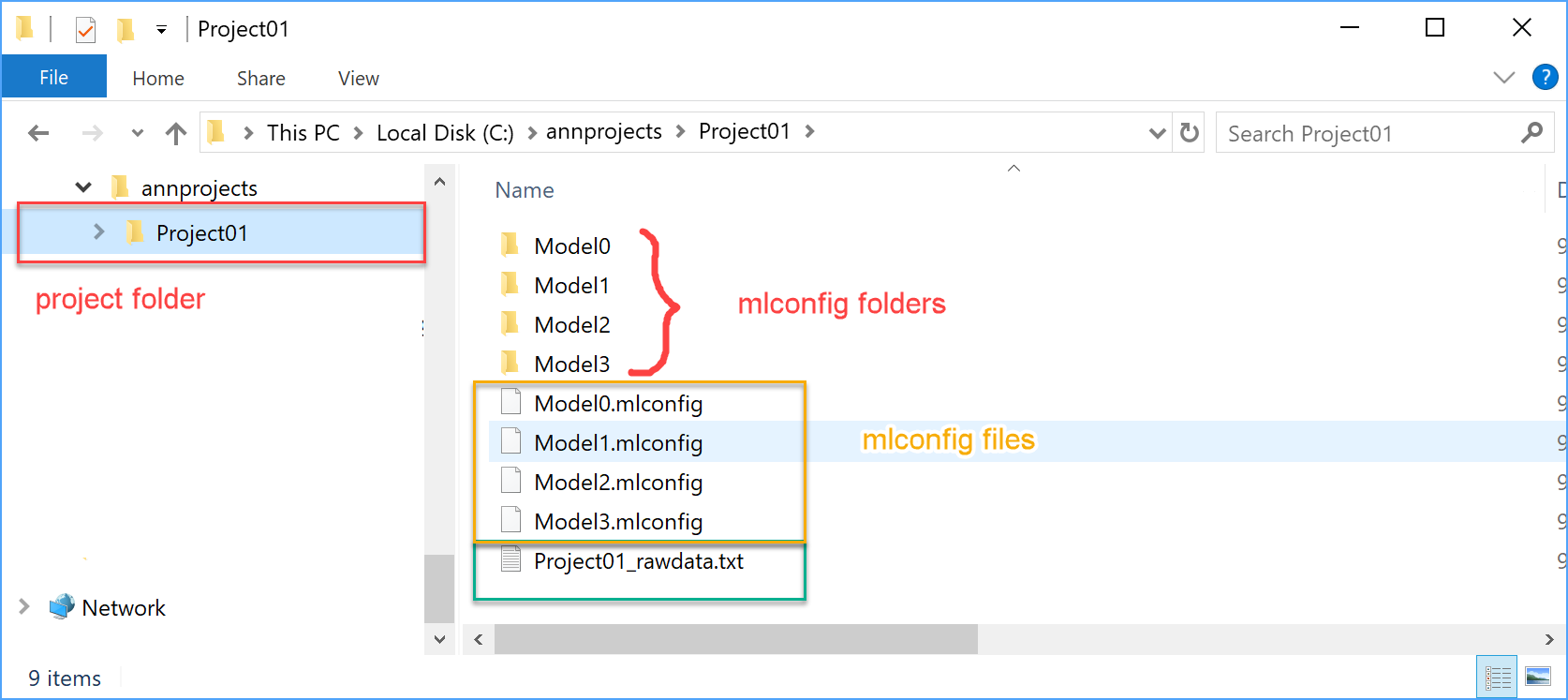}
\caption{Models and related mlconfig files and folders structure}
\label{fig_05} 
\end{figure}

As an example \ref{fig_05} shows the ANNdotNET project with 4 DL models: $Model0$, $Model1$, $Model2$ and $Model3$.

\begin{figure}[h]
\centering
\includegraphics[width=15cm]{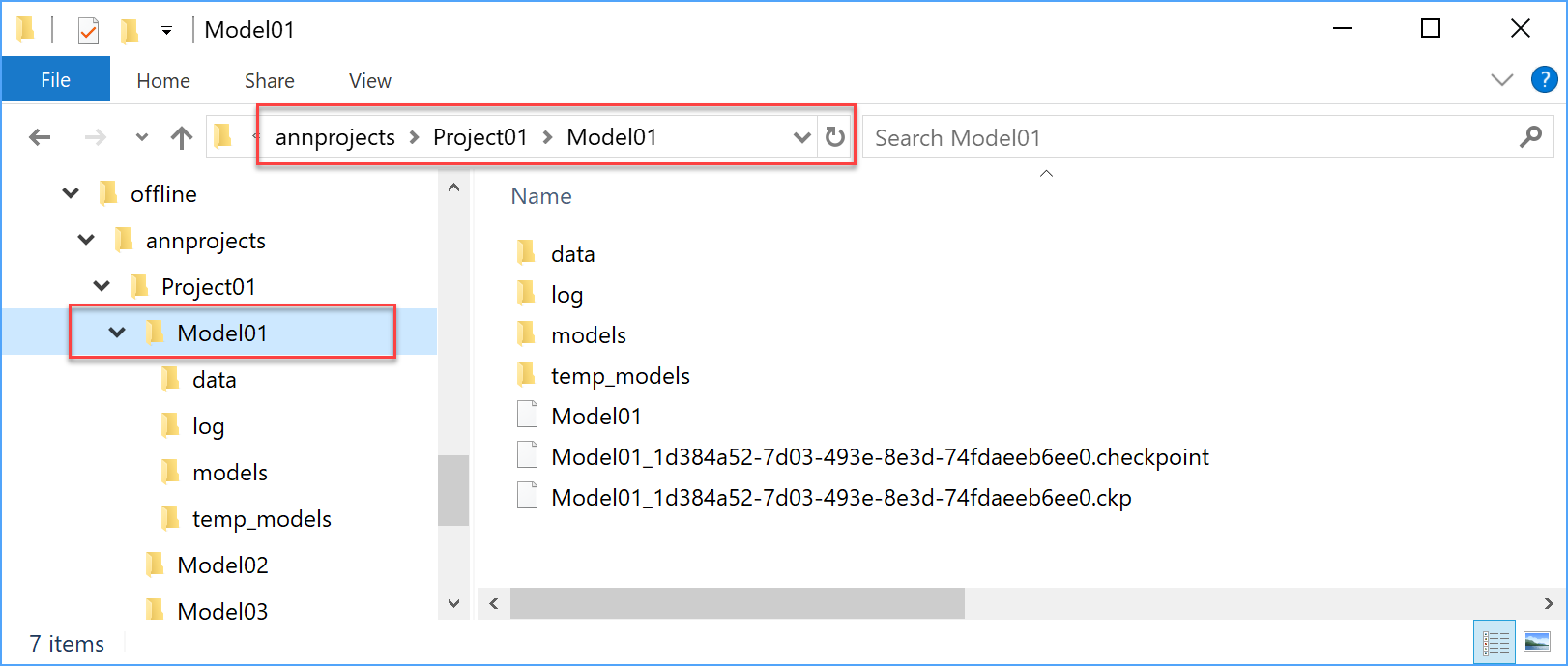}
\caption{Models and related mlconfig files and folders structure}
\label{fig_06} 
\end{figure}

During the models creation separate folder and mlconfig file were created. This kind of file structure offers clean and easy way to follow file structures and information generated in each model (Figure \ref{fig_06}), as well an easy way to transfer mlcofing file to different project. 

Depending of stage of completeness, the model may consist of the following folders and files:

\begin{itemize}
    \item \textbf{data}  contains training, validation and testing ml ready data set,
    \item \textbf{log}  contains files of training information
    \item \textbf{models}  files of CNTK format created during various phase of training
    \item \textbf{temp\_ models}  folder holding temporary model files during training. All content from the folder is deleted once the training process is completed.
  \item  \textbf{model checkpoint state files}  model files stored current the state of the trainer. The files are needed in case when the user want to continue with training based on the previous training state.  
\end{itemize}
	
All model mlconfig file is always placed at the root of the project folder.
 
\subsection{ANNdotNET project file}

ANNDotNET project is stored in \emph{annproject} file. It contains information about data set and DL models. Each project also consists of project info file. It is a Rich Text Format (RTF) file containing the necessary information about the project. The annproject file is text based file consisting of:
\begin{enumerate}
    \item $project$ contains information of the project and related models
    \item $data$ contains information about raw data set.
    \item $parser$ parser information while parsing data set file.   
\end{enumerate}

The $project$ keyword defined the basic project property like:
\begin{itemize}
\item $Name$ name of the project,
\item $ValidationSetCount$  the size of validation data set,
\item $PrecentigeSplit$  is the validation data set size in percentage while creating it,
\item $MLConfigs$ list of created ml configurations,
\item $Info$  project info file.
\end{itemize}

For example, the following text represent typical annproject:

\begin{lstlisting}
!annprojct file for daily solar production

project:|Name:SolarProduction  |ValidationSetCount:20 |PrecentigeSplit:1 
        |MLConfigs:LSTMMLConfig |Info:

!raw dataset and metadata information
data:|RawData:SolarProduction_rawdata.txt |Column01:time;Ignore;Ignore;Ignore; 
                                        |Column02:solar.past;Numeric;Feature;Ignore;
                                        |Column03:solar.current;Numeric;Label;Ignore; 

!parser information 
parser:|RowSeparator:rn |ColumnSeparator: ; |Header:0 |SkipLines:0

\end{lstlisting}

The code above defined the example of the annproject named $SolarProduction$, with raw data set stored in $SolarProduction\_rawdata.txt$ file that contains three columns: $time$, $solar.past$ and $solar.total$. The first column ($time$) is marked as $ignored$ which means it will be excluded from the model training. The $solar.past$ column is marked as feature and $solar.current$ is marked as label. Both feature and label are numeric column. Those information is enough that ANNdotNET tool can created $ml ready datasets$.

The $parser$ keyword is used while the raw datset is loaded into the application memory.

\subsection{\emph{mlconfig} file}

The basic object in ANNdotNET is deep learning model which is represented by the $mlconfig$ file.

The structure of the $mlconfig$ file is described by the 8 keywords:
\begin{itemize}
    \item $configid$ $-$ unique identifier of the \emph{mlconfig} file,
    \item $metadata$ $-$ meta information about data set,
    \item $features$ $-$ defines features for the model,
    \item $labels$ - defines labels for the model,
    \item $network$ - defines neural network architecture to be trained, 
    \item $learning$ - defines learning parameters,
    \item $training$ - defines training parameters,
    \item $path$ – defines paths to files needed during training and evaluation.
\end{itemize}

Each of the above keyword consists of several parameters and values. The syntax of the \emph{mlconfig} file allows you to create as many empty lines as you like. In case you want to add comment in the file, the sentence must begin with exclamation "!". Order of the keywords is irrelevant.

The following content represent typical $mlconfig$ file:

\begin{lstlisting}

!*******ANNdotNET v1.0************
!Iris mlconfig file iris.mlconfig

!configid represent the unique identified of the configuration
modelid:33fe0968-d640-4b53-97dc-982dcf2b1cad

!metada contains information about data set. 
metadata:|Column01:sepal_length;Numeric;Feature;Ignore; 
        |Column02:sepal_width;Numeric;Feature;Ignore;  
        |Column03:petal_length;Numeric;Feature;Ignore; 
        |Column04:petal_width;Numeric;Feature;Ignore;
        |Column05:species;Category;Label;Ignore;setosa;versicolor;virginica 


!Information about features. 
!The line contains two groups of features: NumericFeatures and Product fetaure
features:|NumFeatures  4 0 |Product 10 0 

!Information about label	
labels:|species 3 0

!Network configuration 	
network:|Layer:Normalization 0 0 0 None 0 0 
        |Layer:Dense 5 0 0 ReLU 0 0 
        |Layer:Dense 3 0 0 Softmax 0 0 

!Learning parameter information
learning:|Type:SGDLearner |LRate:0.01 |Momentum:1
        |Loss:CrossEntropyWithSoftmax|Eval:ClassificationAccuracy|L1:0|L2:0
        
!Training parameters information
training:|Type:default |BatchSize:65 |Epochs:1000 |Normalization:0 
        |RandomizeBatch:False |SaveWhileTraining:1 
        |ProgressFrequency:50 |ContinueTraining:0
        |TrainedModel:models\\model\_at\_952of1000\_epochs\_TimeSpan\_636720117054117391 


!Components of the mlconfig paths
paths:|Training:data\mldataset_train 
        |Validation:data\mldataset_valid |Test:data\mldataset\_valid
        |TempModels:temp\_models |Models:models |Result:FFModel\_result.csv |Logs:log 
     
\end{lstlisting}

mlconfig file can be defined using only text editor and then use ANNdotNET  for training and evaluation. Full description of the file can be found at the project repository documentation.

\section {ML Engine - training and evaluation of deep learning models }

ANNdotNET introduces the ANNdotNET Machine Learning Engine (MLEngine) which is responsible for training and model evaluation defined in the mlconfig file. The ML Engine relies on Microsoft Cognitive Toolkit, CNTK open source library for deep learning. Through all application ML Engine exposed all great features of the CNTK e.g. GPU support for training and evaluation, different kind of learners, but also extends CNTK features with more evaluation functions (RMSE, MSE, Classification Accuracy, Coefficient of Determination, etc.), Extended Mini-batch Sources, Trainer and model evaluation.

ML Engine also contains the implementation of neural network layers which supposed to be high level CNTK API very similar as layer implementation in Keras\cite{tensorflow} and other python based deep learning APIs. With this implementation the ANNdotNET implements the Visual Network Designer (VND)  which allows to design neural network configuration of any size with any type of the layers. The following layers are implemented:

\emph{Normalization Layer} – takes the numerical features and normalizes its values at the beginning of the network.
\emph{Dense} – classic neural network layer with activation function
\emph{LSTM} – special version of recurrent network layer with option for peephole and self-stabilization.
\emph{Embedding} – Embedding layer,
\emph{Drop} – drop layer,

Complete list of supported layer can be found in the project documentation.

Designing deep neural networks can be simplify by using pre-defined network layer with capability to created any network we usually implement through the source code.

\subsection{Training and learning parameters}
IN deep learning there are two kind of parameters. The learning and training parameters. The learning parameters are parameters needed during network learning. This includes:

\begin{itemize}
    \item Learner - optimization method used during learning process,
    \item learning rate - the number between 0 and 1 using which determines the step size at each iteration while moving toward a minimum of a loss function.
    \item Loss function used to determine how well learner models the given data,
    \item Evaluation function to measure how close the model predict the output values.
\end{itemize}

ANNDotNET supports mini-batch training which allows to make different type of training strategies.The training parameters includes: number of epochs, mini-batch size and progress frequency. Epochs and mini-batch size are self explanatory. However, the progress frequency is the number of epoch skip until the next epoch is shown in the output. In ANNdotNET the full list of training parameters includes:
\begin{itemize}
    \item Epoch - the number of full cycles when training.
    \item Mini-batch size number of samples in the batch which going into the network,
    \item Progress frequency - shows output of the training progress at every $n$ epoch,
    \item Randomize mini-batch - randomize mini-batch during training process,
    \item Continue training - The parameters indicate if the model will be continue with training, or the training will start from scratch,
    \item Save good models during training - saves model which has better performance parameters than previous one.
\end{itemize}

Training process can be visually monitored by using two graphs:
\begin{itemize}
    \item Mini- batch training - shows the value of the loss and evaluation functions for each mini-batch.
    \item Model evaluation - shows the values of evaluation function for training and validation data set for the current iteration. 
\end{itemize}

\begin{figure}[h]
\centering
\includegraphics[width=15cm]{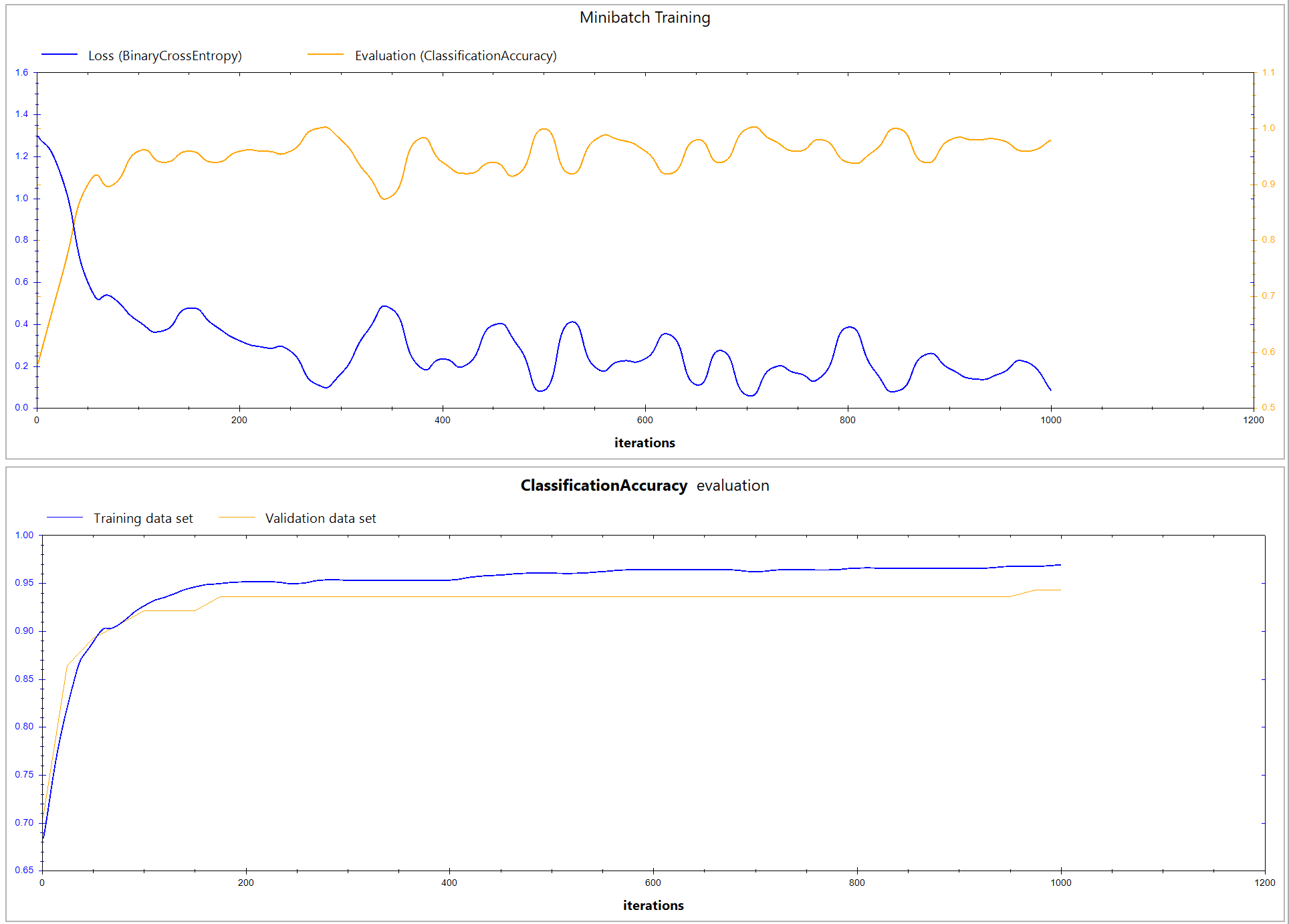}
\caption{Training module in ANNdotNET. During training the training progress is monitored by two diagrams mini-batch training and model evaluation.}
\label{fig_10} 
\end{figure}

The visualization of the training progress can give the user a better picture how training process behave. Is the training process converges, is it going to over-fitted area. The user can stop the training process at any time. Once the training process is stopped or completed the best model is determined based on the selected training strategy (with or without early stopping).

\subsection{Training with early stopping}

In ANNdotNET the early stopping is implemented so that the best trained model is selected after the training process. The best model is selected among other models saved during the training. This is kind of training strategy leads that regardless of the epoch number the best model is always selected without over-fitting.   

\subsection{DL Model evaluation}

\begin{figure}[h]
\centering
\includegraphics[width=15cm]{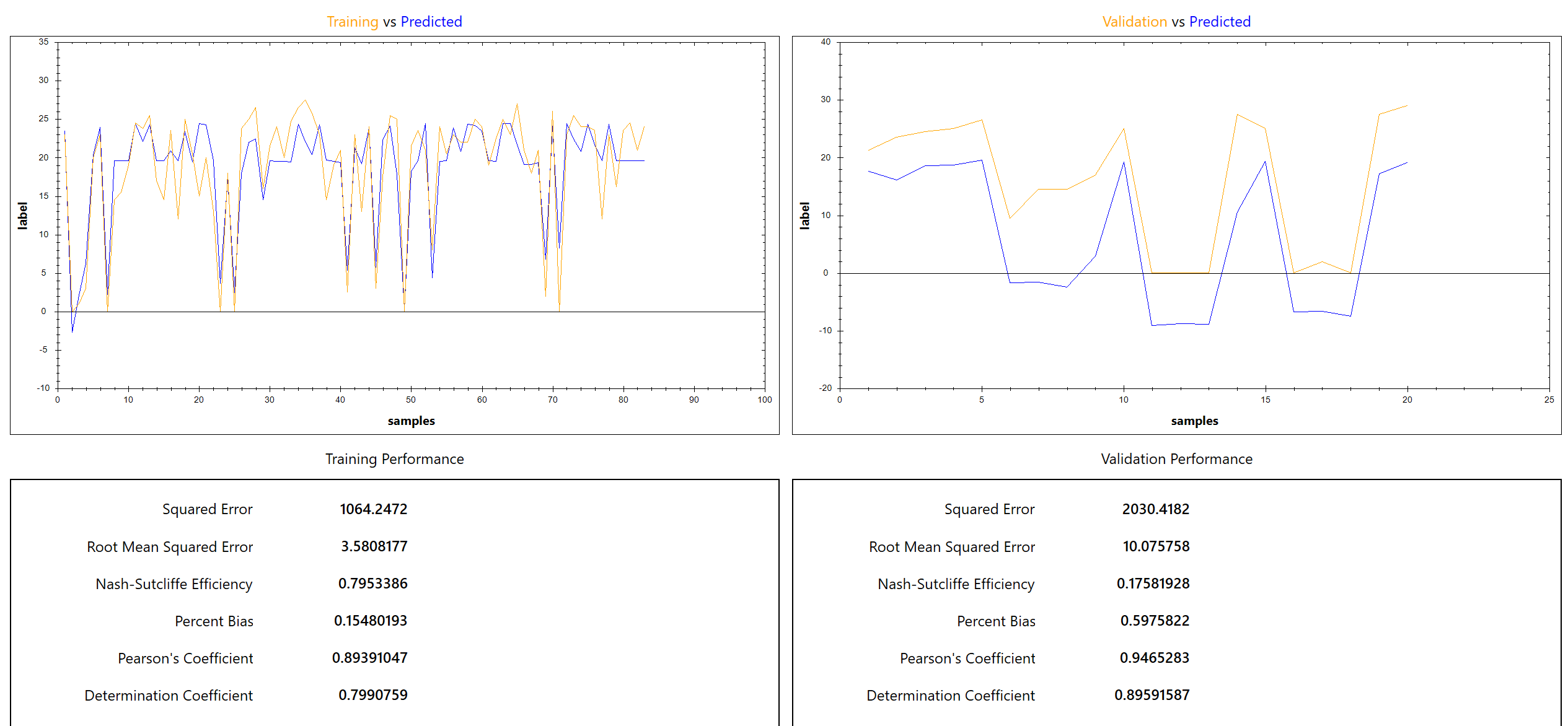}
\caption{Evaluation of regression model}
\label{fig_11} 
\end{figure}

The model evaluation module evaluate the best trained model and presents the performance parameters both for training and validation sets. Depending of the ML type (regression, binary or multi class classification) performance parameters are calculated and presented. Figure \ref{fig_11} shows the evaluation of the regression DL model. However, it is supported both binary and multi class classification model evaluation. 

\section {Visual Network Designer}

Building network is one of the most challenging task in deep learning and it is followed once the set of features and labels are defined. The first layer in the network is the input layer which directly depends of the input data (features). On the other hand, the output layer is defined by the output data (labels). The first and the last layers in the network are defined by the training set while hidden layers are defined with specific architecture. ANNdotNET introduces the Visual Network Designer (VND) which allows to visually create different types of deep network architecture. VND supports basic network layers such dense, dropout, LSTM, convolution and allows to create a network with any combination of the layers. By using proper combination of the basic network layers one can create network such Feed Forward networks, Deep Feed Forward networks, Convolutions Network, Recurrent LSTM based Network, Auto-Encoder, CUDAStackedLSTM network, CUDAStackedGRU network, etc. VND supports creation popular network architecture such AlexNet and similar, or create popular layers such as Autoencoder-Decoder, etc.

Beside classic neural network layers, ANNdotNET implements custom layers such as \emph{Normalization} and \emph{Scale} Layer. Normalization layer normalizes the training data set by calculating the standard deviation and mean of each numeric feature and produces the z-Score as output. With the normalization layer each numeric feature has zero mean and standard deviation of one. This is typical normalization method in training deep leaning models. Scale layer is suitable when normalizing the input data in image recognition tasks.

VND is accessible from the \emph{Network Settings} tab page in the DL Model. The concept of VND is based on sequential list of network layers, so the designer can add, insert remove any available network layer mentioned above. Figure \ref{fig_06} shows an example of CNN network architecture designed to model popular \emph{Cat vs. Dog} data set\cite{catdog}. As can be seen, the figure shows first several network layers sequentially ordered in the list. On the left side one can find the information of data set (training and validation), the  input and the output layer, as well as learning parameters (learning rate, momentum, loss and evaluation functions). In order to change the current network or design new one there are set of options located at the top of the layer list. There are Combo Box with all supported network layers followed by buttons to add, remove and insert network layer. Once the network is designed the \emph{Graph} option can visually represents the while network with its weights, inputs and output parameters.  

\begin{figure}[H]
\centering
\includegraphics[width=15cm]{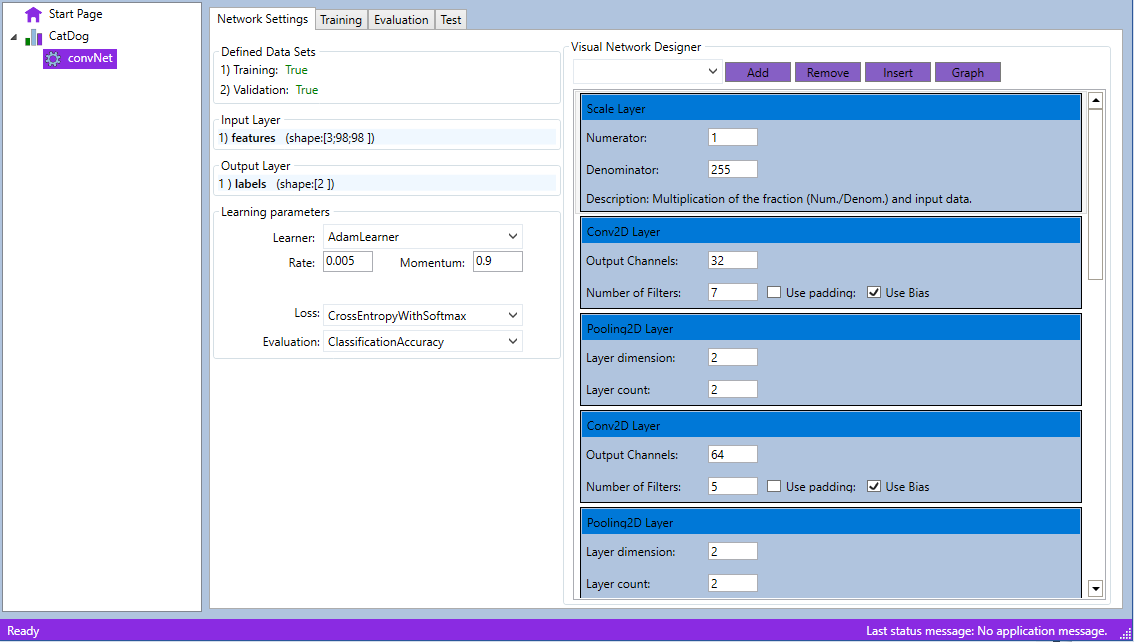}
\caption{Visual Network Designer presenting convolutions network architecture capable to predict Cat/Dog data set}
\label{fig_07} 
\end{figure}

\section{ANNdotNET Excel Addin}

ANNdotNET supports the deployment of the DL model into Microsoft Excel Application by using Excel Add-in. With the Add-in the DL model is used like ordinary formula which can be run from the formula bar. In order to run DL model within Excel, the model should be exported and saved on known location. In Excel, calling the model is achieved by typing the formula:

$$
=ANNdotNET([cell range], [model path])
$$

\begin{figure}[H]
\centering
\includegraphics[width=15cm]{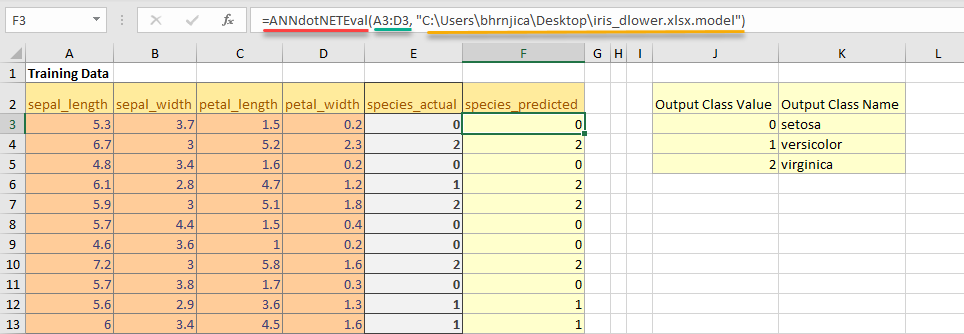}
\caption{Trained model deployed in Excel}
\label{fig_09} 
\end{figure}

Figure above shows Iris model exported in Excel. The predicted values are calculated directly in Excel by calling ANNdotNET funtion within Excel and pointing to cell range and exported model path.

\subsection{ANNdotNET as A Cloud Solution}

By using the ANNdotNET, it is possible to incorporate Deep Learning, (DL) tasks into a cloud solution, so that the complete DL process can be automatized and defined into one workflow using cloud services. 

It can be detected three common tasks in DL cloud solution:

\begin{itemize}
	\item Data preparation 
	\item Training ML model 
	\item Model Deployment 
\end{itemize}

In all three phases ANNdotNET can be incorporated and used.

\begin{figure}[H]
	\centering
	\includegraphics[scale=.65]{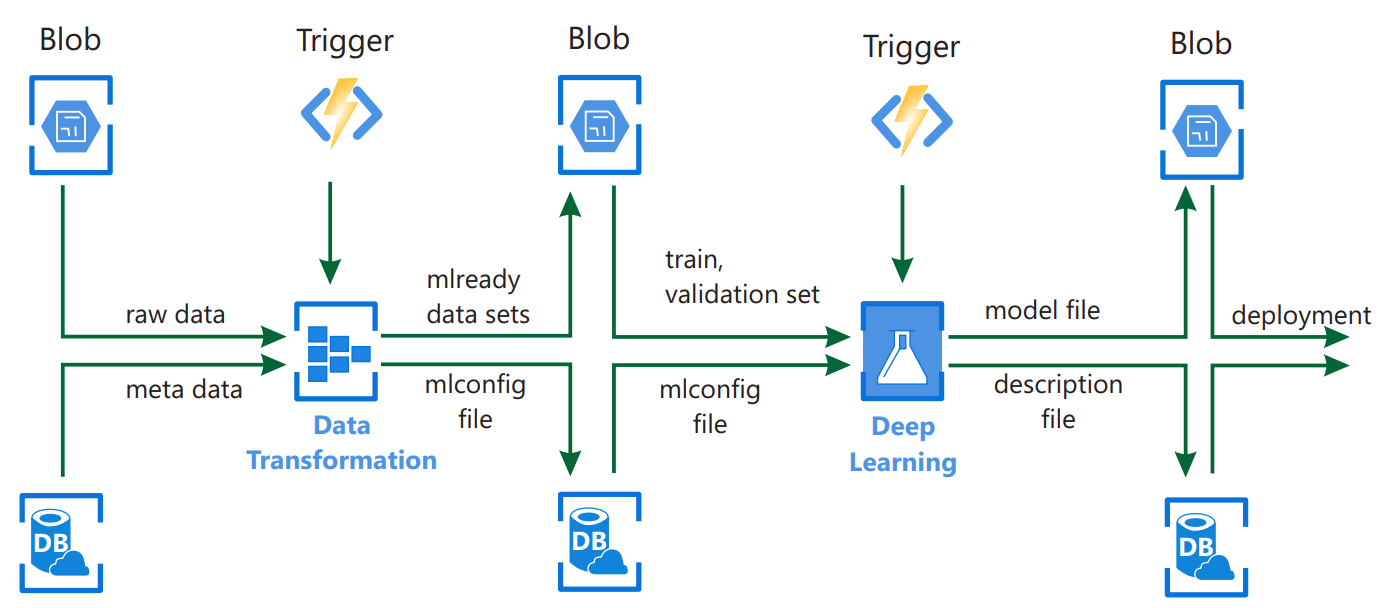}
	\caption{Architecture of Deep Learning Cloud Solution}
	\label{fig_08}       
\end{figure}

The typical scenario can be described on the following:

Once the model configuration is loaded using the mlconfig file, the training process can be started by defining the number of epochs, or by defining the early stopping criteria. The training process can be monitored by reading the training progress information. The information helps the user to decide is the training process converging at the expected speed, or when to stop the training process in order to prevent model over-fitting. The training module that shows training history is shown on Figure \ref{fig_08}. The model deployment is the last phase of the ML cloud solution, and defines several options that can be used for different scenarios. The most common option is to generate a simple web service that contains the implementation of the model evaluation. The web service returns the model output in an appropriate format. The model can also be deployed in Excel, to allow the model to behave as an Excel formula. Excel deployment is achieved by implementing additional Excel add-in. The deployment ML model in Excel is usually suitable when dealing with the input data which is relatively easy to represent in Excel.

The complete cloud ML solution is depicted in Figure \ref{fig_08}. By using the ANNdotNET, it is possible to transform data and prepare it for training. Moreover, ANNdotNET provides components for training, evaluation, testing and deploying deep learning models. Its components can be used in similar cloud solutions depicted in Figure \ref{fig_08}, particularly for "data transformation" and "deep learning" cloud solution components.

\section{Application of ANNdotNET for developing deep learning models}

In this section successful applications are going to be presented. The applications can be classified with the domain problems.

ANNdotNET has been successfully used in many deep learning and ANN research papers and online articles. \cite{hrnjica2019} used ANNdotNET in order to develop LSTM based deep learning models for predicting Vrana lake water level in 6 and 12 months ahead located in Croatia. Authors also used the tool to develop Feed Forward model for comparison results. \cite{zhu2020} used ANNdotNET in order to develop deep learning models to predict lake level for 100 lakes in Poland. Furthermore \cite{hrnjica2} used ANNdotNET in order to develop deep learning model for predicting energy demands in one of the mayor city in Cyprus.

ANNdotNET has been used to develop deep learning model for sentiment analysis \cite{hrnjica2018} and Time Series prediction \cite{hrnjica2017}.  

Beside using recurrent LSTM and feed forward deep networks successful application has been achieved in using convolutions networks mainly for image classifications. Using popular CIFAR-10 \cite{alex2009} data set ANNdotNET achieve prediction with average accuracy higher than 0.96 \cite{hrnjica2018a}. Using Kaggle Cats vs. Dogs data set\cite{catdog} ANNdotNET achieved the accuracy higher than 0.95 which can be found as standard example. Also MNIST \cite{mnist} data set used in order to created deep learning model based on convolutions network with prediction accuracy with more than 0.95. It is also part of the standard package. 
It is also worth mention that ANNdotNET installation package comes with 12 complete and ready to used deep learning project made based on most popular data sets from different problem domains like regressions, binary classifications, multi class classification, time series and image classifications. 

\section{Conclusion}

ANNdotNET is deep learning framework implemented on .NET Framework which is developed for building, training and evaluation of deep learning models. The tool can be used as regular Desktop application providing rich set of user interfaces. The project is completely open sourced and hosted at \url{http://github.com/bhrnjica/anndotnet}. The project is mainly targeting academicians, researchers and hobbies to work with designing deep learning networks. ANNdotNET can also be used through the development environment in order to develop more complex scenario for custom data processing or network design with multiple inputs and outputs or non supported network layers.

\bibliographystyle{unsrt}  


\end{document}